%% file: main.tex
\documentclass[10pt,twocolumn,letterpaper]{article}

\usepackage[pagenumbers]{cvpr} 
\usepackage{multirow}
\usepackage{float}
\usepackage[accsupp]{axessibility}  

\input{preamble}

\definecolor{cvprblue}{rgb}{0.21,0.49,0.74}
\usepackage[pagebackref,breaklinks,colorlinks,allcolors=cvprblue]{hyperref}
\usepackage{bbding}

\def\paperID{15283} 
\def\confName{CVPR}
\def\confYear{2026}

\title{EgoXtreme: A Dataset for Robust Object Pose Estimation \\ in Egocentric Views under Extreme Conditions}

\author{
Taegyoon Yoon$^1$ \quad Yegyu Han$^1$ \quad Seojin Ji$^1$ \quad Jaewoo Park$^1$ \\
Sojeong Kim$^1$ \quad Taein Kwon$^{2*}$ \quad Hyung-Sin Kim$^{1*}$ \\[1ex]
$^1$Seoul National University \qquad $^2$VGG, University of Oxford \\
{\tt\small \{taegyoun88, yegyuhan, seojinji23, 1qkrwodn1, kvia2230, hyungkim\}@snu.ac.kr} \\
{\tt\small taein@robots.ox.ac.uk}
}

\begin{document}

\twocolumn[{%
\renewcommand\twocolumn[1][]{#1}%
 \maketitle
 \begin{center}
     \centering     
     \vspace{-2ex}
     \captionsetup{type=figure}
     \includegraphics[height=7.7cm]{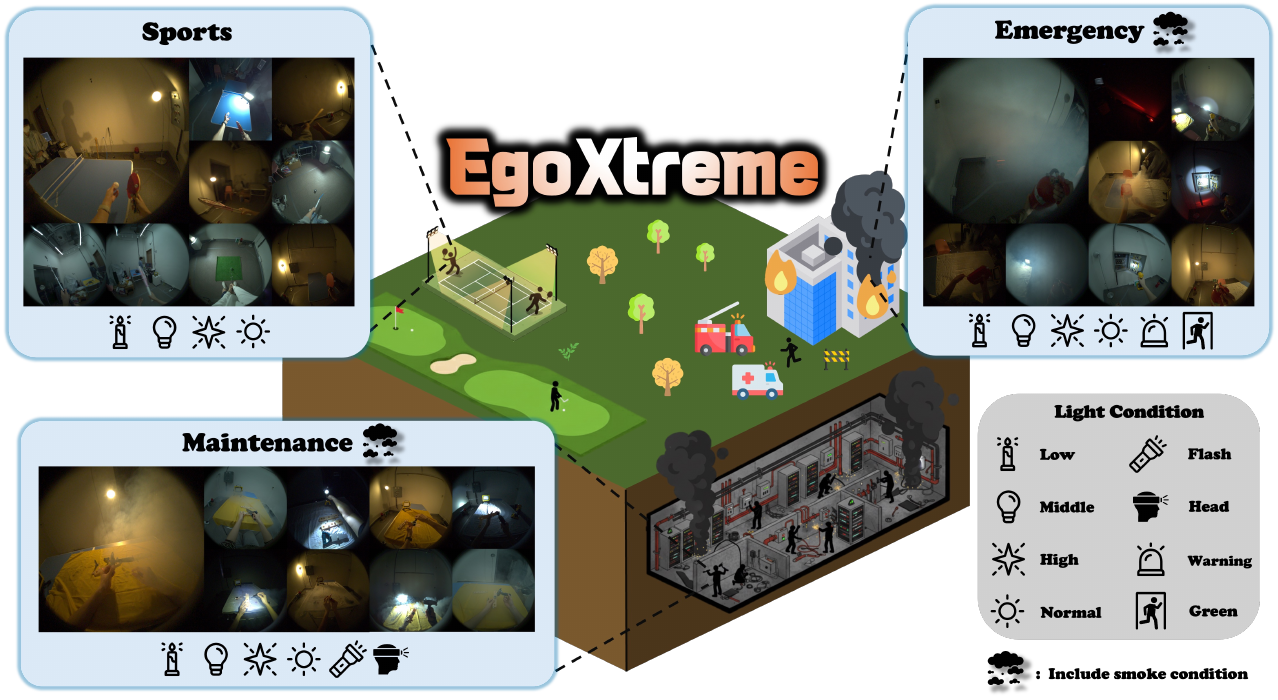}
     \vspace{-2ex}
     \captionof{figure}{\textbf{\ours, an egocentric dataset for robust 6D object pose estimation in extreme environments.} The dataset provides 775.5 minutes of egocentric RGB video from 15 participants using Aria glasses. As illustrated, it spans three challenging scenarios—Sports, Maintenance, and Emergency—featuring significant real-world visual degradations such as low light, smoke, and motion blur.}
 \end{center}%
}]
\renewcommand\thefootnote{}
\footnotetext{*Joint supervision and corresponding authors.}
 \input{sec/0_abstract}    
 \input{sec/1_intro}
 \input{sec/2_relatedwork}
 \input{sec/3_dataset}
 \input{sec/4_experiments}
 \input{sec/5_conclusion}
 {
     \small
     \bibliographystyle{ieeenat_fullname}
     \bibliography{main}
 }

\input{sec/X_suppl}

\end{document}

%% file: preamble.tex
\newcommand{\ours}
{{\textit{EgoXtreme}}\xspace}



%% file: sec/0_abstract.tex
\begin{abstract}

Smart glass is emerging as an useful device since it provides plenty of insights under hands-busy, eyes-on-task situations. To understand the context of the wearer, 6D object pose estimation in egocentric view is becoming essential. However, existing 6D object pose estimation benchmarks fail to capture the challenges of real-world egocentric applications, which are often dominated by severe motion blur, dynamic illumination, and visual obstructions. This discrepancy creates a significant gap between controlled lab data and chaotic real-world application. To bridge this gap, we introduce \ours, a new large-scale 6D pose estimation dataset captured entirely from an egocentric perspective. \ours features three challenging scenarios—industrial maintenance, sports, and emergency rescue—designed to introduce severe perceptual ambiguities through extreme lighting, heavy motion blur, and smoke. 
%
Evaluations of state-of-the-art generalizable pose estimators on \ours indicate that their generalization fails to hold in extreme conditions, especially under low light.
We further demonstrate that simply applying image restoration (e.g., deblurring) offers no positive improvement for extreme conditions. While performance gain has appeared in tracking-based approach, implying using temporal information in fast-motion scenarios is meaningful.
We conclude that \ours is an essential resource for developing and evaluating the next generation of pose estimation models robust enough for real-world egocentric vision.
The dataset and code are available at {\footnotesize\url{https://taegyoun88.github.io/EgoXtreme/}}
\end{abstract}
\vspace{-7ex}

%% file: sec/1_intro.tex
\begin{figure*}
  \centering
  \includegraphics[width=\textwidth]{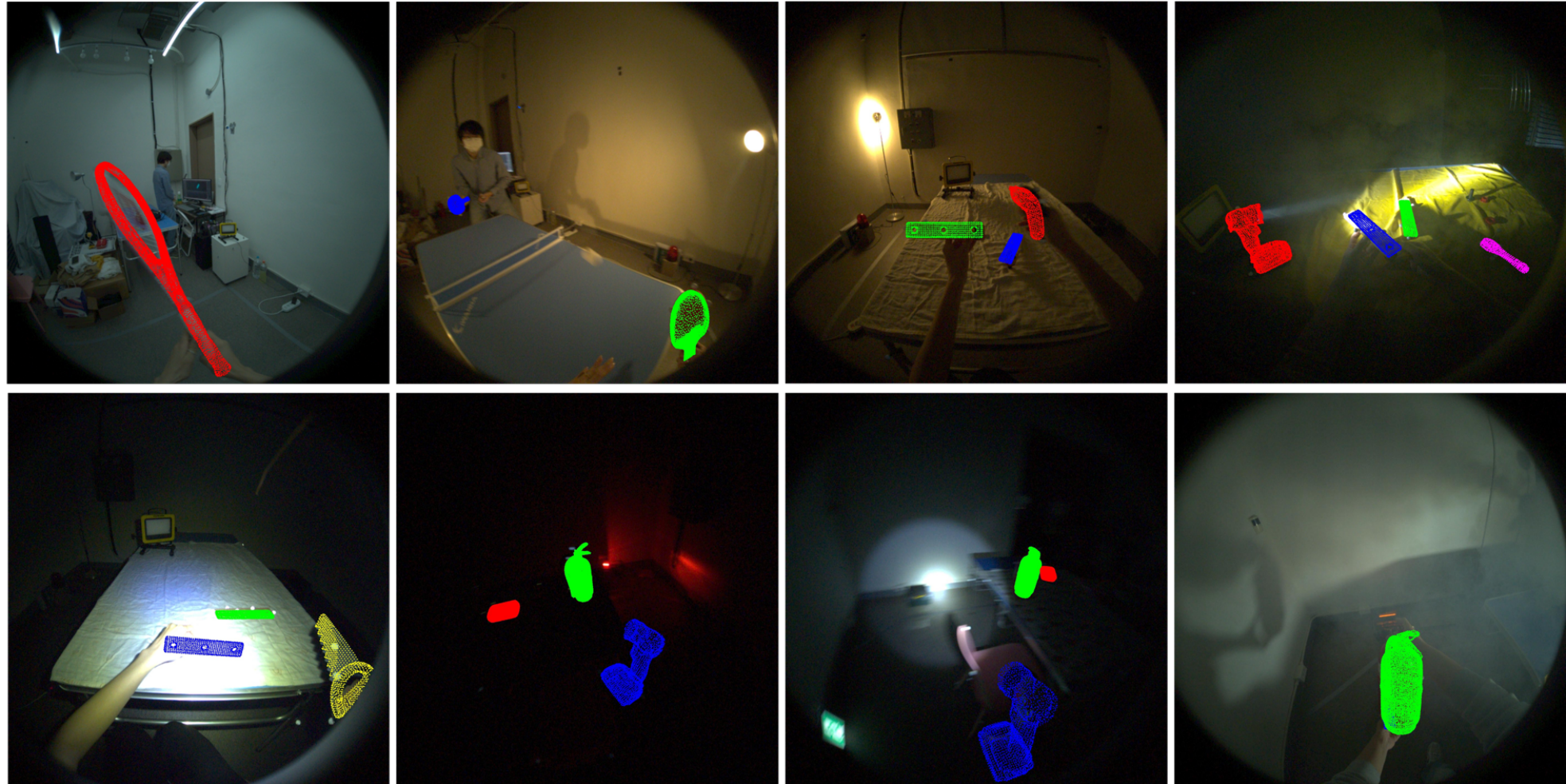}
  \vspace{-4ex}
  \caption{\textbf{Visualization of ground truth 6D pose annotations from our dataset, arranged sequentially from top-left to bottom-right.} The frames include the sports scenario (normal and middle light), the industrial maintenance scenario (low light, flashlight with smoke, and headlight), and the emergency rescue scenario (warning light, exit green light, and high light with smoke). } 
  \vspace{-2ex}
  \label{fig:introduction_visualize}
\end{figure*}

\section{Introduction}
\label{sec:intro}





Lightweight smart glasses~\cite{aria} provide a powerful interface that can continuously capture and interpret the wearer’s \textit{egocentric} view, enabling vision applications precisely when users \textit{cannot free both hands or divert their gaze to a phone}. 
Examples include tightening a bolt while following a torque sequence, navigating a dim corridor to locate an extinguisher, or executing a high‑speed swing. 
In these moments, the camera is rigidly mounted to a moving head; objects appear at close range and are often truncated by hands or tools; and illumination is supplied or modulated by the wearer (headlamp or flash) or by the environment (warning beacons, exit‑sign LEDs).

%
While 6D object pose estimation is fundamental to understanding physical activities and environments---and has achieved strong results on static third‑person RGB datasets~\cite{LM, LMO, tudl, tless, ycbv, hope, hb, icbin, itodd}---egocentric settings introduce unique challenges. 
In \textbf{hands‑busy, eyes‑on‑task, always‑moving} scenarios, the data distribution departs from stationary, uniformly lit third‑person scenes and instead shifts toward rapid and frequent head motion; a near-field perspective at very short working distances; severe truncation at image boundaries; and extreme visual conditions (dynamic or narrow-spectrum illumination, smoke, and clutter) that induce motion blur and visibility loss. The very advantages of smart glasses as an interface therefore create the \textbf{sensing extremes} we must target.

\begin{table*}[h]
\caption{\textbf{Datasets for object pose estimations.}}
\label{tab:dataset_list}
\centering
\resizebox{0.9\textwidth}{!}{
\begin{tabular}{l|cccc|ccccc}
\toprule
& \multicolumn{4}{c|}{\textbf{Extreme Condition}} & \multicolumn{5}{c}{} \\
\cmidrule(lr){2-5}
\textbf{Dataset} &
\textbf{Instance} &
\textbf{Speed} &
\textbf{Light} &
\textbf{Smoke} &
\textbf{Frames} &
\textbf{Egocentric} &
\textbf{Subjects} &
\textbf{Objects} &
\textbf{Annotation} \\
\midrule

LM~\cite{LM} & No & - & 1 & No & 18.2k & No & - & 15 & RGB-D \\
IC-BIN~\cite{ycbv} & Yes & - & 1 & No & 177 & No & - & 2 & RGB-D \\
T-LESS~\cite{tless} & No & - & 1 & No & 49k & No & - & 30 & RGB-D \\
YCB-V~\cite{ycbv} & No & Slow & 1 & No & 0.1M & No & - & 21 & RGB-D \\
TUD-L~\cite{tudl} & No & - & 8 & No & 62k & No & - & 3 & RGB-D \\
HOPE~\cite{hope} & No & - & 5 & No & 238 & No & - & 28 & RGB-D \\
IPD~\cite{ipd} & Yes & - & 3 & No & 30k & No & - & 20 & RGB-D \\
H2O~\cite{h2o} & No & Slow & 1 & No & 572k & Yes & 4 & 8 & RGB-D \\
HOT3D~\cite{hot3d} & No & Slow & 1 & No & 1.5M & Yes & 19 & 33 & Mocap \\

\midrule
\textbf{EgoXtreme (Ours)} & Yes & Fast & 8 & Yes & 1.3M & Yes & 15 & 13 & Mocap \\
\bottomrule
\end{tabular}}
\end{table*}


Despite these realities, the field of 6D object pose estimation largely relies on controlled, third-person datasets such as YCB-Video~\cite{ycbv}, Honnotate~\cite{honnotate}, LineMod~\cite{LM} , T-LESS~\cite{tless}, all collected under stable lighting and limited motion.
While the H2O~\cite{h2o} and HOT3D~\cite{hot3d} introduced egocentric benchmarks, they do not fully capture the diversity and severity of real-world smart-glasses conditions---particularly rapid object manipulations and dynamic lighting.
In such extremes, severe motion blur significantly degrades the generalization of existing models.
Attempts to bridge the gap via simulation---synthetic data generation~\cite{blenderproc}, frame averaging to simulate blur~\cite{realblur}, and artificial brightness variations~\cite{exlpose}---remain insufficient: a substantial discrepancy persists between simulated and real egocentric scenes.
The absence of an egocentric dataset that \textit{directly captures} these extreme conditions is therefore a critical bottleneck for advancing 6D object pose estimation in egocentric vision.

This work aims to step toward bridging the gap between controlled laboratory settings and extreme cases that can occur in real-world usage, analyzing where current pose estimators fail under challenging egocentric conditions. To this end, we introduce \ours, a novel dataset specifically designed to evaluate robustness under dynamic lighting, smoke, and severe motion blur---conditions that induce significant perceptual ambiguity. The dataset comprises video sequences collected from 15 participants over 775.5 minutes, and Figure \ref{fig:introduction_visualize} provides representative sample images from \ours. 

Our dataset presents three distinct scenarios:
\begin{itemize}
\item \textbf{Industrial maintenance:} Fine manipulation using tools like hammers and drills, testing robustness to subtle, precise motions.

\item \textbf{Sports:} Rapidly swung objects (e.g., table tennis rackets, baseball bats), testing resilience to extreme motion blur from high-speed rotational and translational movement.

\item \textbf{Emergency rescue:} Recreating urgent situations, users search for and interact with emergency items (e.g., first-aid kits, extinguishers), testing object detection and tracking amid intense camera shake and visual obstructions.
\end{itemize}

Our extensive evaluation of 6D object pose on \ours reveals that current state-of-the-art pose estimators~\cite{foundationpose,gigapose,picopose}---despite large-scale pretraining aimed at generalizing to unseen objects in cluttered scenes---are highly fragile under egocentric extremes. This indicates that the challenges captured by \ours are largely orthogonal to those addressed by existing generalizable methods, exposing a gap in current benchmarks for egocentric applications. 
Specifically, we identify two primary failure modes: (1) distribution shift induced by dynamic illumination and smoke, and (2) feature loss caused by severe truncation and motion blur. 
Critically, the degradation persists even when perceptual image quality improves: visual restoration (e.g., deblurring, dehazing, and low-light enhancement) offers little benefit, and model performance remains significantly reduced. In contrast, incorporating temporal information via pose tracking improves accuracy in dynamic, high-motion scenarios, underscoring the importance of temporal modeling for 6D pose in egocentric video.


The core contributions of this paper are as follows:
\begin{itemize}
\item \textbf{A large-scale egocentric benchmark for robust pose estimation:} We release \ours, the first large-scale egocentric 6D pose estimation dataset collected across versatile scenarios under extreme conditions, including severe motion blur, dynamic lighting, and visual obstructions. This dataset can serve as a challenging benchmark for smart-glasses applications. 


\item \textbf{Comprehensive baseline analysis:} We evaluate state-of-the-art generalizable 6D object pose estimators on \ours, analyze their limitations and failure modes in extreme egocentric environments, and study two mitigation strategies---image restoration and temporal tracking---demonstrating the critical importance of the latter for robust performance under severe motion.
\end{itemize}

%% file: sec/2_relatedwork.tex
\begin{table*}[t]
    \centering
    \caption{\textbf{\ours datasets configuration}}
    \label{tab:EgoXtreme_configuration}
    \resizebox{0.9\textwidth}{!}{
    \begin{tabular}{l|ccc|ccccc|c|c|cc}
        \toprule
        \multirow{3}{*}{\textbf{Scenario}} & \multicolumn{8}{c|}{\textbf{Lighting condition}} & \multirow{3}{*}{\textbf{Smoke}} & \multirow{3}{*}{\textbf{Object}} & \multicolumn{2}{c}{\textbf{Speed (m/s)}} \\
        \cline{2-9} \cline{12-13}
        & \multicolumn{3}{c|}{\textbf{Standard}} & \multicolumn{5}{c|}{\textbf{Extreme}} & & & \multirow{2}{*}{\textbf{Camera}} & \multirow{2}{*}{\textbf{Object}} \\
        \cline{2-9}
        & normal  & middle & high & low & head & flash & warning & green & & \\
        \midrule
        Maintenance & \CheckmarkBold & \CheckmarkBold & \CheckmarkBold & \CheckmarkBold & \CheckmarkBold & \CheckmarkBold & & & \CheckmarkBold & 5 & 0.03 & 0.09 \\
        Sports & \CheckmarkBold & \CheckmarkBold & \CheckmarkBold & \CheckmarkBold & & & & & & 5 & 0.40 & 1.37 \\
        Emergency & \CheckmarkBold & \CheckmarkBold & \CheckmarkBold & \CheckmarkBold & & & \CheckmarkBold & \CheckmarkBold & \CheckmarkBold & 3 & 0.47 & 0.10 \\
        \bottomrule
    \end{tabular}}
\end{table*}

\section{Related work}
\label{sec:relatedwork}

This section reviews existing research pertinent to our work, covering 6D object pose estimation models, relevant benchmark datasets, and methodologies aimed at improving performance under adverse conditions such as blur, haze, and low light.

\subsection{Benchmarks for 6D object pose estimation}
Standard benchmarks like YCB-Video~\cite{ycbv} and LM/LM-O~\cite{LM, LMO} advanced the field but they only controlled limited conditions like lighting and motion. Datasets focusing on specific challenges like multiple instances/symmetry (IC-BIN~\cite{icbin}, IC-MI~\cite{icmi}, T-LESS~\cite{tless}) or static lighting variations (TUD-L~\cite{tudl}, HOPE~\cite{hope}, SenseShift6D~\cite{ss6d}) exist, but lack the dynamic or extreme lighting. Table \ref{tab:dataset_list} indicates the difference between \ours and the other 6D pose estimation benchmarks. 

6D pose estimation methods are broadly divided into instance-specific models (e.g., GDRN~\cite{gdrnet}, ZebraPose~\cite{zebrapose}, HiPose~\cite{hipose}) and zero-shot models. Instance-specific models are impractical for our egocentric scenario as they require object-specific retraining for novel objects. While zero-shot models offer generalization, RGB-D approaches (e.g., SAM6D~\cite{sam6d}, FoundationPose~\cite{foundationpose}) are often incompatible with lightweight smart glasses that lack depth sensors. We therefore focus on RGB-only zero-shot models. Two main strategies exist within this category. The first one is a two-step pipeline where a coarse estimator like GigaPose~\cite{gigapose} and  FoundPose~\cite{foundpose} retrieves template candidates via feature matching for P$n$P-RANSAC~\cite{pnp, ransac}, and a separate refinement module (e.g., MegaPose~\cite{megapose}, Gen-Flow~\cite{genflow}) iteratively improves the best hypothesis by aligning dense features. The second strategy uses integrated models like PicoPose~\cite{picopose}, which perform the entire pipeline internally: they first find a single best-matched template and then use dedicated multi-stage blocks to compute 2D transformations and refine the final pose. Since our egocentric scenario requires an RGB-only unseen approach, we analyze the robustness of these state-of-the-art coarse-to-fine pipelines on our proposed extreme environment benchmark.

\subsection{Egocentric pose estimation datasets}
Recently, datasets capturing the first-person perspective have become crucial for smart glass applications. However, a significant portion of this research has centered on human-centric aspects, such as full-body 3D pose (e.g., You2Me~\cite{you2me}, EgoBody~\cite{egobody}) or detailed hand pose and hand-object interactions (e.g., H2O~\cite{h2o}), rather than the 6D pose of general objects in the environment. HOT3D~\cite{hot3d}, the most prominent large-scale egocentric object tracking benchmark, was a significant step in addressing this. However, it was captured under relatively well-lit conditions and similarly lacks the extreme motion, challenging illumination, or visual obstructions like smoke that we target. Consequently, a significant gap remains in evaluating object pose estimation robustness under the severe motion blur, dynamic/extreme lighting, and drastic movements typical of real-world smart glass usage.

\subsection{Methodologies for adverse conditions}
Various methods aim to enhance image quality under adverse conditions. Deblurring techniques range from GAN-based approaches~\cite{deblurganv2} to recent Transformer~\cite{restormer} and efficient architectures~\cite{nafnet}. Dehazing methods include classic approaches~\cite{dcp} and deep learning models~\cite{ffanet} or Transformer-based~\cite{dehazeformer}. Low-light enhancement research offers specialized solutions, ranging from efficient zero-reference methods to multi-task restoration models. For instance, Zero-DCE~\cite{zerodce} formulates enhancement as a fast, image-specific curve estimation task, while DarkIR~\cite{darkir} jointly addresses low-light, noise, and blurring issues by employing advanced attention mechanisms within efficient CNNs. While potentially beneficial for detection/segmentation, the impact of these restoration methods and potential artifacts on downstream pose estimation accuracy, especially in extreme conditions, is not well-studied. We investigate the effectiveness of applying representative enhancement techniques as preprocessing for robust 6D pose estimation on our benchmark.

%% file: sec/3_dataset.tex
\section{\ours dataset}
\label{sec:dataset}

\begin{figure}
  \centering
  \includegraphics[width=\linewidth]{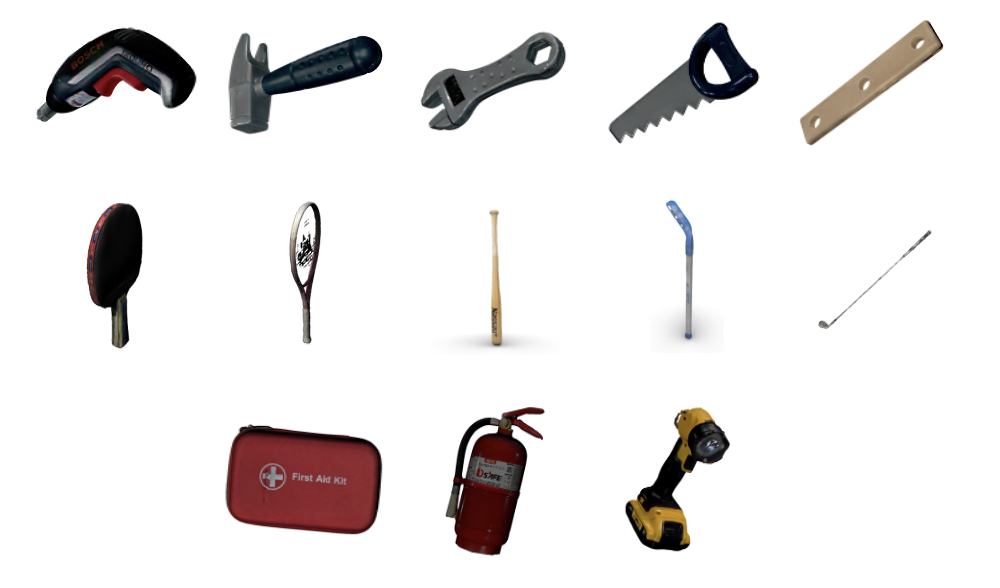}
  \caption{\textbf{3D models.} This image shows the 13 object models in the \ours dataset. From top to bottom five maintenance scenarios, five sports scenarios and three emergency scenarios.}
  \label{fig:dataset_objects}
\end{figure}

\ours is a novel, large-scale dataset designed for robust egocentric 6D object pose estimation under extreme conditions. Specifically, 8 illumination conditions are used across three scenarios, and smoke is included in specific scenes. The configurations of whole scenario is summarized in Table~\ref{tab:EgoXtreme_configuration}. 
These conditions, combined with severe motion blur, make accurate 6D object pose estimation extremely challenging.

Comprising approximately 1.3 million frames (775.5 minutes total) captured at 30fps (1650 frames/55 seconds per sequence), the dataset is split into training (518.8 min), validation (80.7 min), and test (176 min) sets. The dataset includes three challenging scenarios: industrial maintenance (319 min), emergency rescue (165 min), and sports (291.5 min). Videos were recorded using Aria glasses~\cite{aria}, providing raw $1408 \times 1408$ fisheye RGB images and the corresponding undistorted version, enabling evaluation on both data types. 15 participants were recorded performing diverse actions and object manipulations. The dataset features 13 objects, including sports equipment, assembly blocks, and emergency supplies, with corresponding 3D CAD models as in Figure~\ref{fig:dataset_objects}. Some scenarios involve multiple instances of the same object (assembly blocks and pingpong racket), requiring instance-level disambiguation.

\begin{figure}
  \centering
  \includegraphics[width=\linewidth]{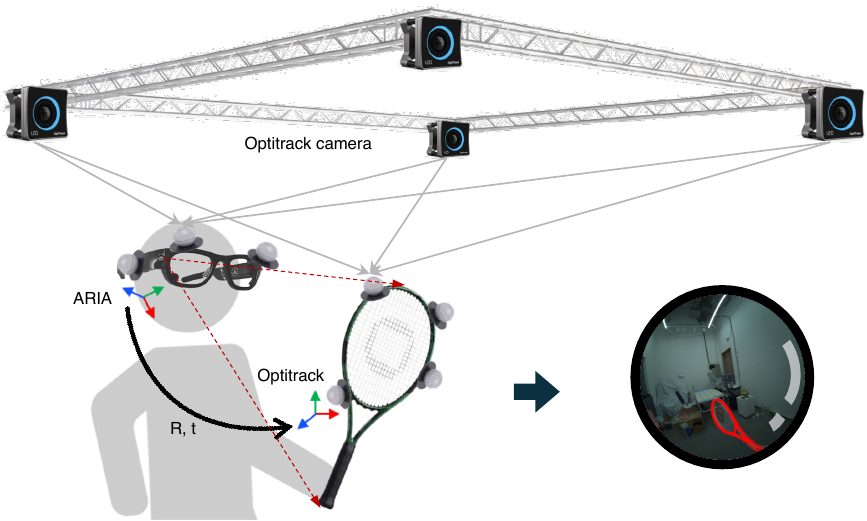}
  \caption{\textbf{Diagram for data collection.}}
  \label{fig:dataset_collection}
\end{figure}

\subsection{Scenario details}
As shown in Table~\ref{tab:EgoXtreme_configuration}, the average speeds of the camera and the object differ significantly depending on the scenario which demonstrates the characteristics of each scenario.

\noindent\textbf{Industrial maintenance scenario} focuses on estimating object pose during seated precision tasks involving manipulating tools, including hammer, drill, saw, wrench and brick, and assembling identical multiple blocks in various industrial lighting conditions. 6 lighting conditions include normal brightness, low light, medium brightness, high brightness, headlamp, and flashlight. To mimic hazy workplaces, smoke was introduced in a subset of the data. This scenario allows for benchmarking model robustness against fine-grained object movements and diverse illumination changes, particularly the movable light sources are included.

\noindent\textbf{Emergency rescue scenario} involves searching for and picking up three types of emergency supplies, such as first aid kit, extinguisher, and flashlight, while moving through different lighting environments. 6 lighting conditions comprise normal brightness, low light, medium brightness, high brightness, emergency beacon, and exit sign lighting. Smoke was added in a portion  to simulate fire situations. This setup evaluates the ability to detect objects and estimate object poses under intense camera shake and restricted visibility.

\noindent\textbf{Sports scenario} features five types of sports equipment, including baseball bat, hockey stick, pingpong racket, tennis racket, and golf club, being swung rapidly. To make it more realistic, actual pingpong gameplay or self-practice are included. 4 lighting conditions are used in this scenario: normal brightness, low light, medium brightness, and high brightness. 
This scenario evaluates the model’s ability to handle extreme motion blur caused by high-speed rotational and translational object movements.

\subsection{Data collection and processing}
RGB images were captured at 30fps using the Aria glasses, synchronized with 1000fps SLAM data. A reflective marker was attached to the Aria glasses, and rigid body marker clusters were attached to all scenario objects, allowing simultaneous recording with a 120fps OptiTrack motion capture system. This system provides sub-millimeter accuracy for the 6D pose of both the headset and the objects, serving as our ground truth (GT). In this setup, both the SLAM trajectory and the motion capture marker trajectory were collected concurrently. The trajectories were aligned using the Umeyama method~\cite{umeyama} to unify coordinate frames and achieve precise temporal synchronization between the Aria SLAM and the motion capture system. The whole process is illustrated in Figure~\ref{fig:dataset_collection}.

To compensate for SLAM drift that occurred during motion, a Kalman filter~\cite{kalman} was applied to refine the trajectory alignment. Once the coordinate frames were unified through this process, various objects tracked by the motion capture system could be represented in the same global coordinate system, enabling accurate object pose annotation. Finally, to correct minor temporal misalignments, on the order of a few milliseconds, between the synchronized Aria timestamp and the true physical capture time of the RGB image caused by the RGB camera’s exposure delay, a manual time-offset adjustment was performed based on visual inspection of projection results before final interpolation.

\begin{table*}[t]
\caption{\textbf{6D object pose estimation on \ours.} Performance is evaluated using ADD(S) recall at thresholds of $0.1d$, $0.2d$, $0.3d$ ($\uparrow$) and spatial accuracy metrics MSSD/MSPD ($\downarrow$), where $d$ denotes the object diameter.}
\label{tab:experiment1}
\centering
\resizebox{\textwidth}{!}{
\begin{tabular}{l|cc| ccccc| ccccc| ccccc}
\toprule
\multirow{2}{*}{\textbf{Scenario}} &
\multirow{2}{*}{\textbf{Light}} &
\multirow{2}{*}{\textbf{Smoke}} &
\multicolumn{5}{c|}{\textbf{FoundPose}~\cite{foundpose}} &
\multicolumn{5}{c|}{\textbf{GigaPose}~\cite{gigapose}} &
\multicolumn{5}{c}{\textbf{PicoPose}~\cite{picopose}} \\
\cline{4-18}
 &  &  &
 \textbf{0.1d} &
 \textbf{0.2d} &
 \textbf{0.3d} &
 \textbf{MSSD} &
 \textbf{MSPD} &
 \textbf{0.1d} &
 \textbf{0.2d} &
 \textbf{0.3d} &
 \textbf{MSSD} &
 \textbf{MSPD} &
 \textbf{0.1d} &
 \textbf{0.2d} &
 \textbf{0.3d} &
 \textbf{MSSD} &
 \textbf{MSPD} \\
\midrule
  \multirow{2}{*}{Sports} & Standard & &
   0.53 & 1.55 & 4.72 & 0.59 & 7.65 &
   4.12 & 11.77 & 24.64 & 5.21 & 10.61 &
   3.13 & 9.48 & 24.61 & 5.00 & 12.87 \\
   & Extreme & & 
   0.18 & 0.78 & 2.42 & 0.23 & 6.87 &
   3.11 & 9.19 & 19.04 & 4.15 & 8.31 &
   1.80 & 6.61 & 17.86 & 3.07 & 10.10 \\
\midrule
  \multirow{4}{*}{Maintenance} & Standard & & 
   21.02 & 30.53 & 37.61 & 12.94 & 18.78 &
   33.64 & 48.84 & 62.77 & 15.97 & 24.17 &
   39.27 & 62.42 & 76.84 & 24.76 & 33.20 \\
   & Extreme & & 
   13.78 & 22.94 & 30.03 & 9.92 & 16.40 &
   19.78 & 32.20 & 45.52 & 10.44 & 19.89 &
   26.44 & 47.18 & 64.09 & 18.37 & 28.12 \\
   & Standard & \CheckmarkBold & 
   14.44 & 22.49 & 30.00 & 8.23 & 13.11 &
   23.01 & 37.87 & 52.86 & 12.24 & 19.83 &
   26.37 & 46.05 & 59.87 & 18.26 & 25.48 \\
   & Extreme & \CheckmarkBold & 
   11.19 & 18.57 & 25.63 & 7.01 & 13.07 &
   17.56 & 30.35 & 45.11 & 8.92 & 18.38 &
   20.97 & 38.50 & 52.30 & 14.25 & 25.53 \\
\midrule
  \multirow{4}{*}{Emergency} & Standard & & 
   6.31 & 11.96 & 12.88 & 9.73 & 13.08 &
   22.03 & 40.29 & 46.34 & 32.08 & 43.52 &
   22.67 & 59.11 & 67.83 & 46.40 & 50.81 \\
   & Extreme & & 
   0.10 & 0.29 & 0.56 & 8.27 & 9.61 &
   9.40 & 14.75 & 21.30 & 12.44 & 29.02 &
   9.18 & 27.59 & 36.23 & 22.01 & 31.24 \\
   & Standard & \CheckmarkBold & 
   3.52 & 9.62 & 12.12 & 0.21 & 1.66 &
   16.25 & 35.28 & 44.91 & 28.34 & 35.99 &
   19.66 & 61.35 & 72.82 & 48.10 & 50.64 \\
   & Extreme & \CheckmarkBold & 
   0.11 & 0.60 & 0.76 & 0.44 & 2.05 &
   7.07 & 15.26 & 21.54 & 12.00 & 32.81 &
   9.45 & 24.19 & 31.54 & 19.37 & 29.62 \\
\bottomrule
\end{tabular}}
\end{table*}

\subsection{Dataset configuration}
The lighting and environmental conditions within the dataset are structured for multi-dimensional evaluation of model robustness, as detailed in Table \ref{tab:EgoXtreme_configuration}. The conditions are classified into two main categories: \textbf{Standard} (common illumination) and \textbf{Extreme} (limited illumination), comprising a total of 8 unique lighting types. The Standard category includes environments with consistent, ample light sources such as normal fluorescent light (normal), bright lamp light (middle), and floodlight illumination (high). Conversely, the extreme category features setups designed to challenge pose estimation: low-intensity lighting (low), dynamic illumination coupled with head movement (head, flash), and critical emergency sources (warning - rotating beacons and green - emergency exit signs). Furthermore, to accurately model visual degradation in maintenance and emergency scenarios, simulated smoke was intentionally introduced using a fog machine. 
This smoke was dispersed to maximize atmospheric scattering and visual obstruction, ensuring that objects were not contaminated directly, thereby isolating the effects of airborne particulates on visibility.


%% file: sec/4_experiments.tex
\section{Experiments}
\label{sec:experiments}

In this section, we evaluate the three state-of-the-art RGB-based models and analyze the performance of 6D object pose estimation under the proposed extreme conditions. Since real-world scenarios demand high generalizability, we focus our evaluation on state-of-the-art zero-shot 6D object pose estimation models. Specifically, we define the testing pipelines for each as follows: FoundPose~\cite{foundpose} is used as a coarse alignment module with its output refined by MegaPose~\cite{megapose}. GigaPose~\cite{gigapose} also utilize the MegaPose as a refinement module. 
Conversely, PicoPose~\cite{picopose} is employed as an integrated coarse-to-fine model performing its own internal refinement. For evaluation, we explicitly utilize the official test set. 
First, we establish baselines on the \ours dataset using state-of-the-art 6D object pose estimation models (Sec.~\ref{sec:poseestimation}). Second, we investigate how image restoration affects 6D object pose estimation performance (Sec.~\ref{sec:withpreprocessing}). Finally, we benchmark and evaluate tracking of the object pose under severe motion blur (Sec.~\ref{sec:tracking}).

\begin{figure*}[t]
  \centering
  \includegraphics[width=\textwidth]{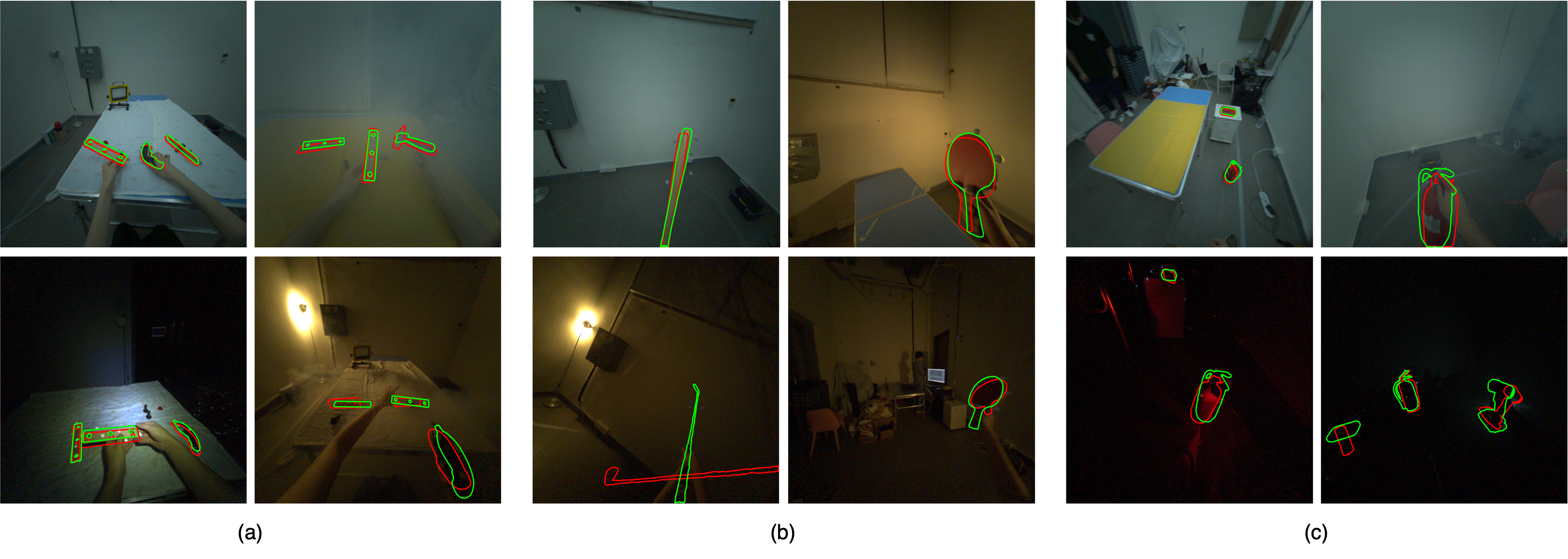}
  \vspace{-5ex}
  \caption{\textbf{Example 6D Pose estimation results on baseline models.} The red line is prediction and green is GT. (a), (b), and (c) are the industry maintenance, sports, and emergency rescue scenarios, respectively. The top row indicates standard light condition, and the bottom row indicates extreme light condition.}
  \vspace{-2ex}
  \label{fig:experiment_qualitative1}
\end{figure*}

\subsection{Baseline evaluation}
\label{sec:poseestimation}

We first establish a baseline with recent state-of-the-art RGB-only zero-shot models, FoundPose~\cite{foundpose}, GigaPose~\cite{gigapose} and PicoPose~\cite{picopose}, on the three scenarios of the \ours dataset. To rigorously evaluate the pose estimation performance decoupled from detection errors, we utilized GT bounding boxes. For evaluation metrics, we employ the ADD(-S)~\cite{LMO, ycbv} recall, along with the standard BOP metrics~\cite{bop}: Maximum Symmetry-Aware Surface Distance (MSSD) and Maximum Symmetry-Aware Projection Distance (MSPD). End-to-end evaluation results using CNOS~\cite{cnos} detections are provided in Appendix C1.

As shown in Table~\ref{tab:experiment1}, the results demonstrate a significant performance degradation in extreme lighting and smoke conditions compared to standard light without smoke conditions. Specifically, PicoPose showed a $31.6\%p$ performance degradation in the emergency scenario (measured @$0.3d$) when comparing the extreme lighting condition to the standard lighting condition. Furthermore, in the industrial maintenance scenario, GigaPose's performance degraded by $9.91\%p$ with the addition of smoke. The sports scenario characterized by severe object cropping and motion blur proved to be the most challenging with models failing to achieve meaningful recall overall. FoundPose exhibited the lowest recall, which we attribute to the brittleness of its direct sparse feature matching approach. Feature extraction and matching under egocentric cropping and motion blur frequently failed to secure the minimum number of correspondences required for PnP-RANSAC~\cite{pnp, ransac}, resulting in a high rate of failure where no 6D pose output was generated. These baselines show that current state-of-the-art models are highly vulnerable to real-world challenges, particularly motion blur and extreme lighting that are not captured in existing datasets. It highlights the necessity of the \ours dataset in filling the gap caused by these real-world challenges. Figure~\ref{fig:experiment_qualitative1} visualizes pose estimation results in standard and extreme cases, respectively.

\subsection{Object pose estimation with pre-processing}
\label{sec:withpreprocessing}

\begin{table}[t]
\caption{\textbf{6D object pose estimation with pre-processing for PicoPose}.}
\vspace{-1ex}
\label{tab:experiment2}
\centering
\resizebox{\linewidth}{!}{
\begin{tabular}{l|ccc| ccc}
\toprule
\multirow{2}{*}{\textbf{Scenario}} &
\multirow{2}{*}{\textbf{Deblur}} &
\multirow{2}{*}{\textbf{Dehaze}} &
\multirow{2}{*}{\textbf{Light enhance}} &
\multicolumn{3}{c}{\textbf{PicoPose}} \\
\cline{5-7}
 & & & &
 \textbf{0.1d} &
 \textbf{0.2d} &
 \textbf{0.3d} \\
\midrule
  \multirow{4}{*}{Sports} & & & & 2.81 & 8.80 & 23.02 \\
  & \CheckmarkBold & & & 2.61 & 8.52 & 22.24 \\
  & & & \CheckmarkBold & 2.87 & 9.05 & 22.05 \\
  & \CheckmarkBold & & \CheckmarkBold & 2.58 & 8.22 & 20.99 \\
  \midrule
  \multirow{5}{*}{Maintenance} & & & & 28.22 & 48.51 & 63.32 \\
  & \CheckmarkBold & & & 26.28 & 43.70 & 57.55 \\
  & & \CheckmarkBold & & 24.04 & 42.75 & 57.71 \\
  & & & \CheckmarkBold & 24.90 & 43.38 & 58.08 \\
  & \CheckmarkBold & & \CheckmarkBold & 23.39 & 39.73 & 53.28\\
  \midrule
  \multirow{5}{*}{Emergency} & & & & 15.12 & 42.69 & 51.70 \\
  & \CheckmarkBold & & & 14.13 & 37.74 & 45.51 \\
  & & \CheckmarkBold & & 4.74 & 22.49 & 37.74 \\
  & & & \CheckmarkBold & 13.78 & 38.51 & 46.75 \\
  & \CheckmarkBold & & \CheckmarkBold & 13.72 & 35.58 & 43.65 \\
\bottomrule
\end{tabular}}
\end{table}

We investigate whether image restoration techniques can mitigate the performance drop under adverse conditions such as motion blur, smoke, and dynamic lighting. To this end, we apply representative preprocessing methods for deblurring~\cite{nafnet}, dehazing~\cite{dehazeformer}, and low-light enhancement~\cite{darkir} directly to the images before running the baseline pose estimator.


The recall values under full lighting and smoke conditions are detailed in Table~\ref{tab:experiment2}. The result of image restoration techniques ultimately fail to help, and often hurt performance. Applying a single preprocessing method generally did not increase performance, and combining two methods resulted in an even greater performance drop, decreasing recall by approximately $5\%p$ (measured @$0.3d$) threshold in the maintenance scenario and $8\%p$ in the emergency scenario. 

\begin{figure}[t]
  \centering
  \includegraphics[width=0.48\textwidth]{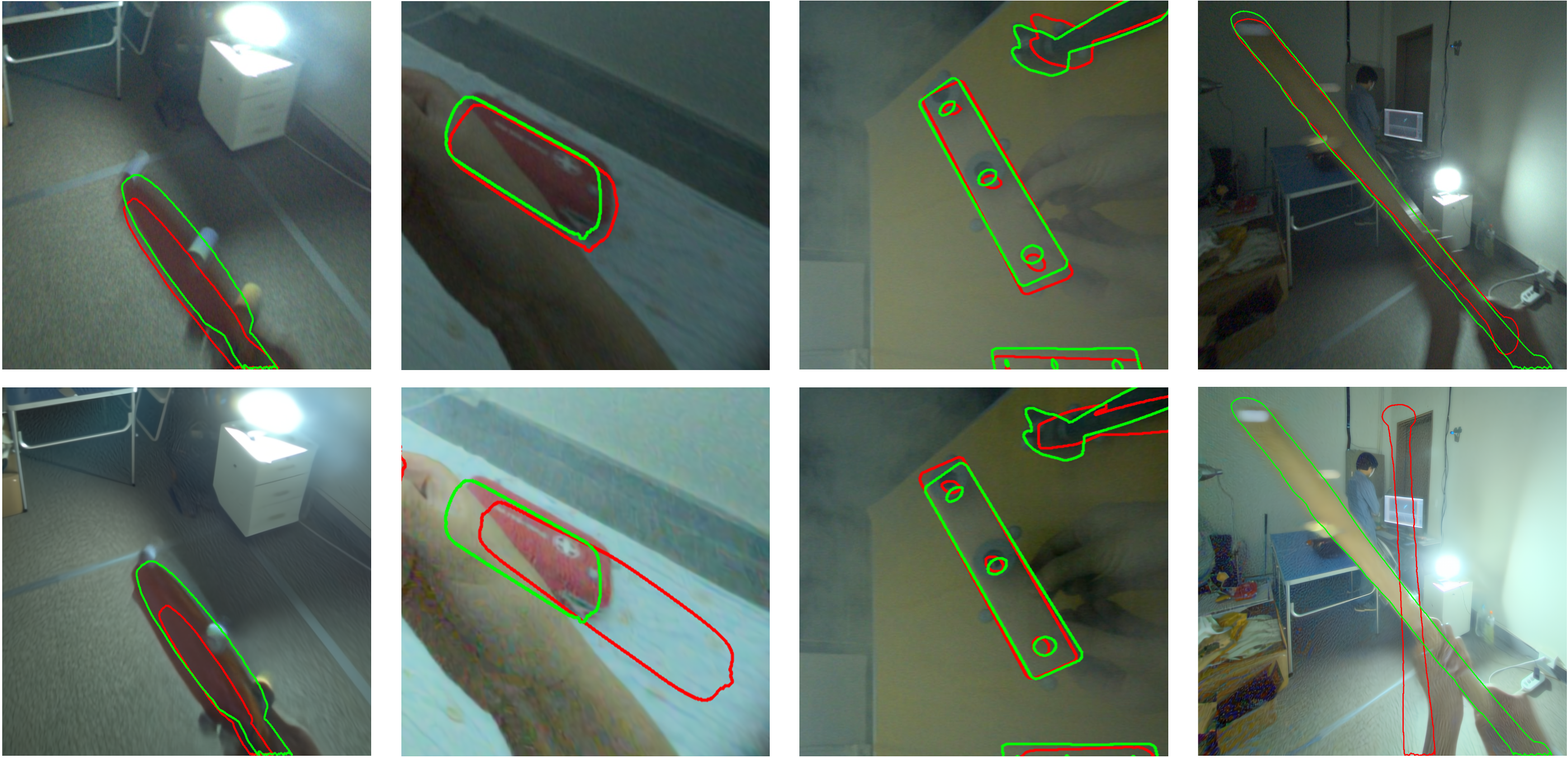}
  \vspace{-4ex}
  \caption{\textbf{Example 6D Pose estimation results with preprocessing.} The top row shows the original, non-preprocessed images. The bottom row displays the corresponding images after applying specific preprocessing: deblurring (left), light enhancement (middle), and dehazing (right). }
  \label{fig:experiment_qualitative2}
\end{figure}

Specifically, when light enhancement faces extreme conditions such as high-contrast highlights or extremely low light, these methods introduced significant noise making prediction more difficult. The current deblurring and dehazing methods also showed no meaningful positive effect. In particular, dehazing method has severely failed in the emergency scenario yielding an exceptionally low recall of only $4.74\%p$ (measured @$0.1d$). The current dehazing method introduces significant noise artifacts under non-uniform smoky conditions. This explains why our emergency scenario characterized by partial smoke was not properly dehazed, making subsequent pose estimation extremely difficult. Although human observers may perceive an improvement in the preprocessed images of some specific scenarios, the model performance decreased as illustrated in Figure~\ref{fig:experiment_qualitative2}.

This result demonstrates that existing preprocessing methods are insufficient for 6D object pose estimation on our dataset and its extreme conditions.
This failure underscores that the challenges posed by \ours cannot be resolved through preprocessing alone and require deeper investigation using our dataset and benchmark.

\subsection{Object pose tracking}
\label{sec:tracking}

\begin{table}[t]
\caption{\textbf{6D object pose tracking for GigaPose}. Applied to sports normal scenario.}
\label{tab:experiment3}
\centering
\resizebox{0.82\linewidth}{!}{
\begin{tabular}{l|c| ccc}
\toprule
\multirow{2}{*}{\textbf{Object}} &
\multirow{2}{*}{\textbf{Method}} &
\multicolumn{3}{c}{\textbf{GigaPose}} \\
\cline{3-5}
 & & \textbf{0.1d} &
     \textbf{0.2d} &
     \textbf{0.3d} \\
\midrule
  \multirow{4}{*}{Pingpong} & Per-frame & 0.53 & 2.97 & 17.63 \\
   & Direct & 0.05 & 0.35 & 6.58 \\
   & Fusion & 0.56 & 1.57 & 10.81 \\
   & Hybrid & 0.49 & 1.52 & 16.53 \\
   \midrule
  \multirow{4}{*}{Tennis} & Per-frame & 6.77 & 34.71 & 50.91 \\
   & Direct & 3.91 & 14.28 & 22.77 \\
   & Fusion & 5.47 & 30.64 & 49.56 \\
   & Hybrid & 6.93 & 34.34 & 50.55 \\
  \midrule
  \multirow{4}{*}{Bat} & Per-frame & 17.95 & 41.29 & 60.55 \\
   & Direct & 4.66 & 9.84 & 14.29 \\
   & Fusion & 13.68 & 37.10 & 60.59 \\
   & Hybrid & 17.66 & 45.27 & 64.46 \\
  \midrule
  \multirow{4}{*}{Golf} & Per-frame & 0.08 & 1.64 & 8.36 \\
   & Direct & 0.64 & 1.49 & 3.61 \\
   & Fusion & 0.45 & 5.76 & 13.98 \\
   & Hybrid & 0.29 & 4.08 & 14.35 \\
   \midrule
  \multirow{4}{*}{Hockey} & Per-frame & 0.46 & 5.66 & 18.35 \\
   & Direct & 0.07 & 3.18 & 5.66 \\
   & Fusion & 1.57 & 12.31 & 24.47 \\
   & Hybrid & 1.08 & 10.72 & 26.13 \\
\bottomrule
\end{tabular}}
\end{table}

For object pose tracking, we compare the tracking-based approach against per-frame baseline (Sec.~\ref{sec:poseestimation}).
We focus on the highly dynamic sports scenarios. We evaluated three distinct temporal strategies: (1) \textbf{Direct temporal} (using the full pose of $t-1$ frame as initial input), (2) \textbf{Fusion temporal} (combining final rotation pose of $t-1$ frame with coarse translation pose of $t$ frame), and (3) \textbf{Confidence-based hybrid temporal} (selectively using the temporal prior based on the current frame's prediction score). 

However, the introduction of the tracking strategy did not yield consistent performance gains across all temporal strategies (Table~\ref{tab:experiment3}), the dynamic nature of the sports scenario clearly revealed the inherent limitations of direct temporal approaches on our dataset. Specifically, the fast object speed in the sports scenario caused large inter-frame displacement. Therefore, the simple direct temporal propagation is inadequate and sometimes adverse to performance, evidenced by a performance degradation of up to $46\%p$ (measured @$0.3d$). Furthermore, the fusion temporal approach was observed to perform worse than the per-frame baseline in some cases (e.g., pingpong and tennis). This failure is due to propagating the previous frame's rotation, which leads to inaccurate initial poses that the current frame measurement struggles to correct. Conversely, the hybrid temporal strategy successfully mitigated these failures, significantly enhancing performance in this high-motion environment. This robustness stems from its ability to selectively ignore unstable rotational priors by utilizing the current measurement's prediction score, thereby maintaining tracking robustness even under severe displacement.

These results strongly suggest that solving pose estimation in extreme dynamic environments requires not merely simple temporal propagation, but a hybrid temporal strategy that uses dynamic quality assessment to selectively apply the temporal prior. 
This inherent challenge, explicitly highlighted by the comparative analysis on our dynamic scenarios, confirms the utility of \ours as a crucial and necessary benchmark for future research. It serves not only to evaluate conventional per-frame estimation accuracy, but, more importantly, to drive the development of robust and resilient pose tracking algorithms capable of operating reliably under real-world extreme conditions.


%% file: sec/5_conclusion.tex
\section{Conclusion}
\label{sec:conclusion}

In this paper, we filled the critical gap between existing 6D object pose benchmarks and the demanding conditions of real-world egocentric vision. We introduced \ours, a new large-scale 6D pose dataset featuring three challenging scenarios specifically designed to test model robustness against severe motion blur, dynamic/low lighting, and visual obstructions like smoke.
Our extensive evaluation of state-of-the-art RGB-only models revealed their significant performance degradation on \ours, confirming that existing methods lack robustness to these extreme conditions. Our experiments showed that while image restoration partially enhances visual fidelity, this improvement fails to translate into meaningful gains for 6D object pose accuracy. Moreover, we show that incorporating temporal information through a tracking-based approach improves pose accuracy and stability in the sports scenario, where motions are extremely fast. To enable performance improvements relevant to these demanding real-world scenarios, a proper benchmark is required for an accurate evaluation, and \ours dataset is specifically introduced to serve this vital role. 

\noindent\textbf{Limitations and future works.} While \ours provides high-fidelity 6D poses for objects, our methodology involves a few trade-offs. Firstly, the reliance on the OptiTrack system for high-accuracy GT confines data collection to specialized indoor environments, limiting the evaluation of true outdoor robustness. 
Secondly, we note the absence of 3D hand pose annotations. Accurately labeling intricate hand articulations, particularly due to the extreme motion blur and speed prevalent in the sports scenario, remains a significant independent challenge. We believe, however, that addressing these gaps presents valuable directions for future work, especially by generating accurate 3D hand labels through combining our existing motion capture data with advanced parametric hand models.

Our findings suggest that future research on robust egocentric object pose estimation should move beyond single-frame visual feature representation and focus on effective temporal modeling and robustness against extreme conditions to handle real-world scenarios. We believe \ours will serve as a critical benchmark to drive the development of the next generation of 6D pose estimation models.

\section*{Acknowledgment}
\label{ack}
This work was supported in part by an SNSF Postdoc. Mobility Fellowship (P500PT\_225450), the Institute of Information \& communications Technology Planning \& Evaluation (IITP) grant funded by the Korea government (MSIT) (No. RS-2023-00216821), and the National Research Foundation (NRF) of Korea grant funded by the Korea government (MSIT) (No. RS-2023-00222663). We thank Meta for providing the Project Aria glasses used in this research. We also gratefully acknowledge the following colleagues for valuable discussions and support of our project: Keondo Park, Eunsu Baek, Joopyo Hong, Yoojin Kwon, Subeom Park, Wooseok Lee, Hun Heo, Seojun Heo, Hongjun Suh, Suahn Bae, Dayeon Woo, Yejun Ji, Wonjeong Lee, Dongik Park, and Boyeong Im. \\

%% file: sec/X_suppl.tex
\clearpage
\setcounter{page}{1}
\maketitlesupplementary
\appendix

\setcounter{figure}{0}
\renewcommand{\thefigure}{A\arabic{figure}}

\section{Dataset specifications}

\noindent\textbf{Capture device} The primary data streams were captured using Project Aria glasses~\cite{aria}, a lightweight platform designed for egocentric machine perception and providing tightly calibrated and time-synchronized streams crucial for our highly dynamic data capture. The core visual stream utilizes one rolling-shutter RGB camera, recording at 30 fps with $1408 \times 1408$ px resolution and an F-Theta fisheye lens ($110^\circ$ FOV). Concurrently, dual integrated Inertial Measurement Units (IMUs) capture motion data at a high frequency (up to 1000 Hz), which is essential for precise temporal synchronization and SLAM trajectory reconstruction. To facilitate precise time synchronization and trajectory alignment with the external motion capture system, seven reflective markers were rigidly attached to the Aria glasses.

\begin{figure}[h]
  \centering
  \includegraphics[width=1.0\linewidth]{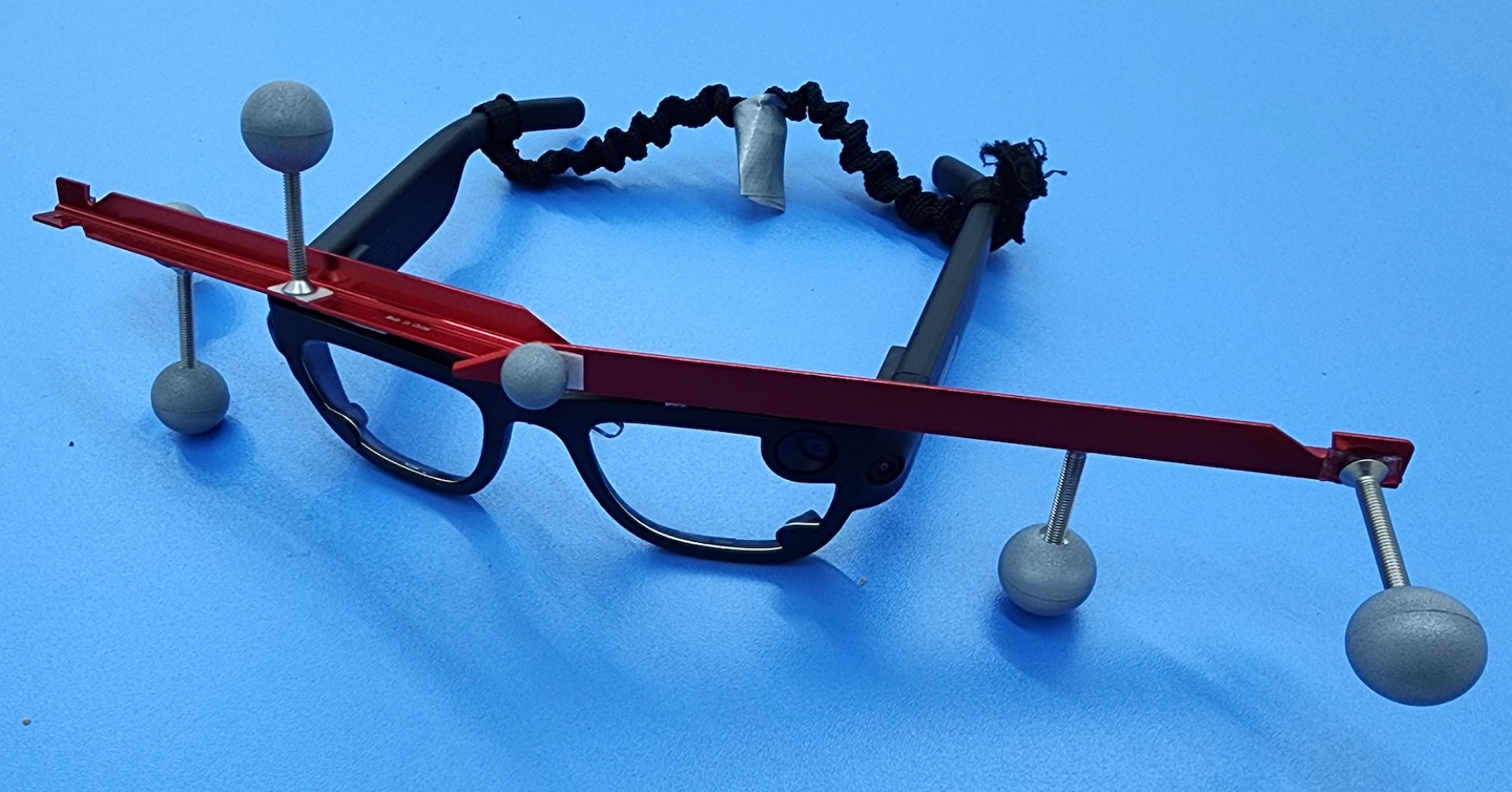}
  \caption{\textbf{Project Aria.} RGB capture device.}
  \label{fig:appendix_aria}
\end{figure}

\noindent\textbf{Test bed} The experiments were conducted within a dedicated laboratory space measuring $2.4 \text{m} \times 2.6 \text{m}$ (width $\times$ depth), with a ceiling height of $3.2 \text{m}$. The room was completely blacked out to eliminate external light interference, ensuring strict control over illumination conditions during testing. High-accuracy ground truth pose data was captured using a motion capture system consisting of four OptiTrack cameras mounted at the $3.2 \text{m}$ height of the ceiling. This system offers high precision, featuring $1280 \times 1024$ resolution, $0.2 \text{mm}$ 3D accuracy, a field of view of $56^\circ \times 45^\circ$, and supports frame rates up to $240 \text{fps}$, ensuring robust tracking even under rapid motion.

\begin{figure}[h]
  \centering
  \includegraphics[width=0.9\linewidth]{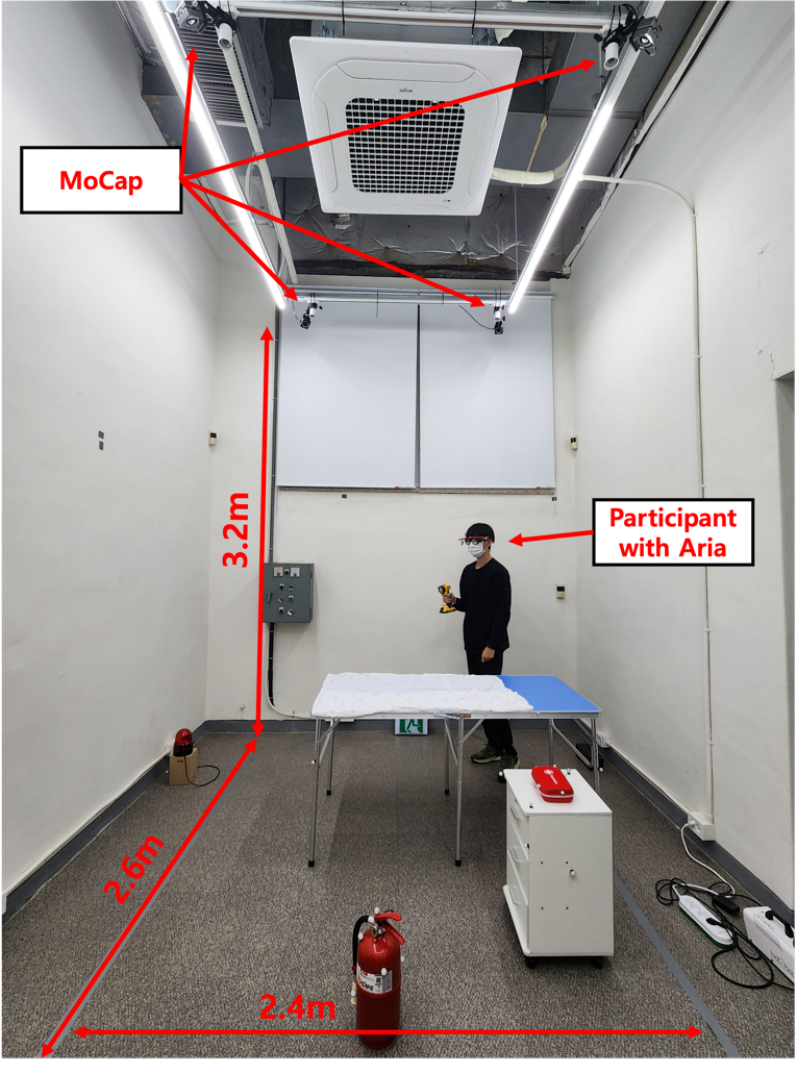}
  \caption{\textbf{Test bed.} Motion capture systems.}
  \label{fig:appendix_testbed}
\end{figure}

\noindent\textbf{Ground Truth Validation} We aligned SLAM and Mocap trajectories to leverage their complementary strengths: while SLAM provides continuous tracking, it is susceptible to low light and fast motion; Mocap (IR-based) remains robust in such conditions but suffers from occasional tracking loss due to occlusion. To validate this alignment, we randomly sampled 200 frames per scenario, selecting only those where keypoints were visibly verifiable. The resulting mean reprojection errors ($1408 \times 1408$ res.) were 12.40 px ($0.62\%$) for sports, 10.77 px ($0.54\%$) for maintenance, and 19.08 px ($0.96\%$) for emergency. The final average trajectory alignment error was $4.3 \text{mm} / 1.5^{\circ}$, validating the reliability of our GT even under such extreme constraints.

\noindent\textbf{Data collection procedure} A total of 15 participants took part in the study. Before commencing data capture, all subjects signed a consent form, and were thoroughly briefed by the principal investigator regarding the study's scope, motion capture boundaries, and specific behavioral guidelines. Crucially, participants were also required to wear protective masks during some sessions involving smoke injection to ensure safety and compliance with protocol. The data collection process required approximately three hours per participant, during which an average of 80 video sequences were captured per subject. In total, 845 video sequences were recorded.

\noindent\textbf{Symmetry properties} We identified \textit{Bat}, \textit{Brick}, and \textit{Tennis} as symmetric objects within our dataset. Consequently, we utilized the ADD-S metric for these objects to account for their geometric symmetry during evaluation, ensuring a robust performance assessment.

\section{Qualitative results of temporal dynamics}

\setcounter{figure}{0}
\renewcommand{\thefigure}{B\arabic{figure}}

\begin{figure}[h]
  \centering
  \includegraphics[width=1.0\linewidth]{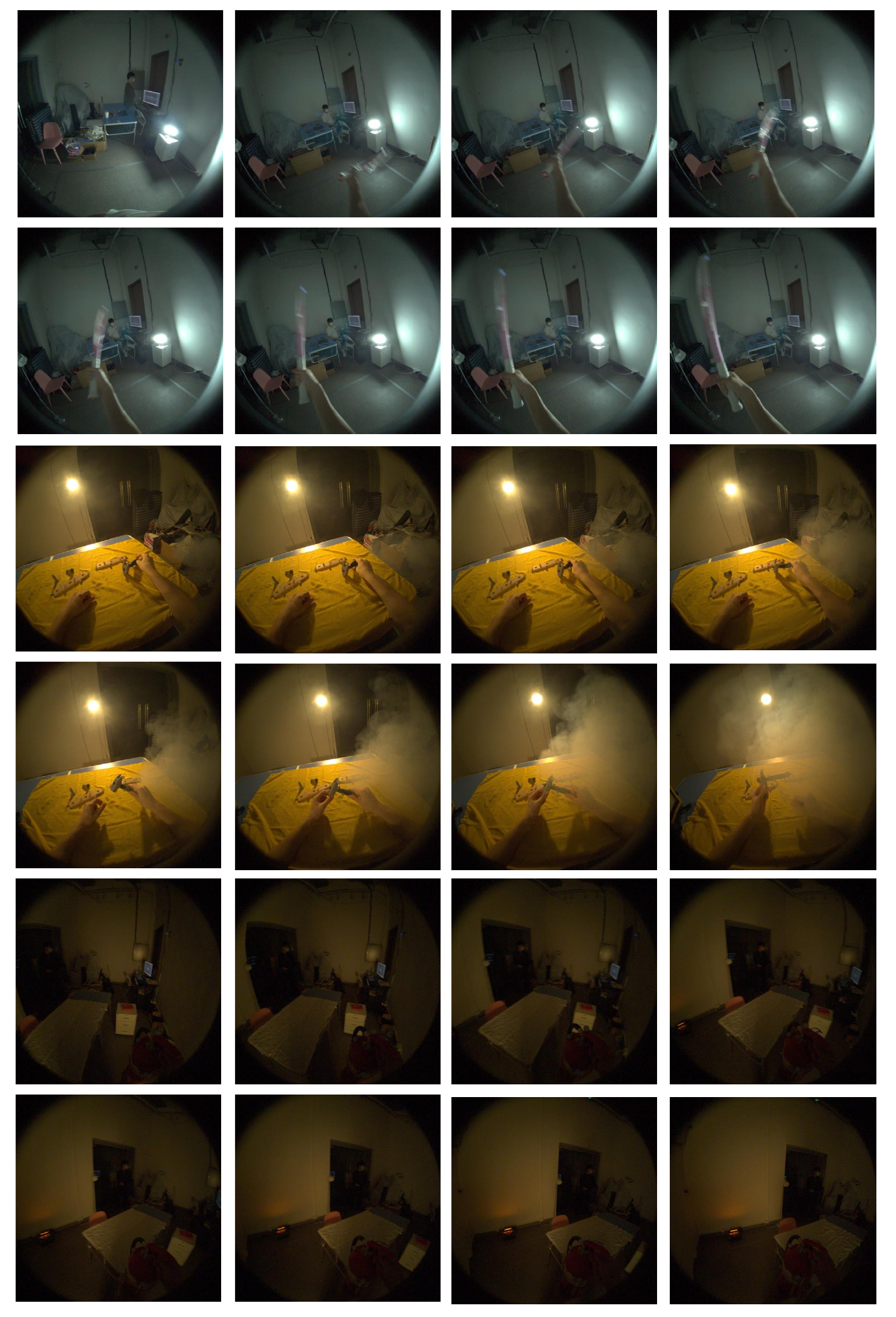}
  \caption{\textbf{Visualization of temporal dynamics in the \ours dataset.}}

  \label{fig:appendix_sample_image}
\end{figure}

Figure~\ref{fig:appendix_sample_image} illustrates the diverse temporal dynamics within our dataset. The visualization is divided into three sections (top-to-bottom), illustrating the dataset’s core challenges and temporal characteristics: rows 1–2 showcase severe motion blur (1-frame intervals); rows 3–4 capture the progression of visual obstruction caused by simulated smoke (10-frame intervals); and rows 5–6 demonstrate rapid egocentric perspective change (4-frame intervals).

\section{Extended baseline analysis}

\subsection{End-to-end 6D object pose estimation}

In Table~\ref{tab:appendix_experiment1}, we present the end-to-end evaluation results using CNOS~\cite{cnos} detections. As observed, the relatively low detection success (AP@0.5) leads to a significant drop in end-to-end pose accuracy (ADD-0.3d) compared to the GT baseline reported in the main paper (Table~\ref{tab:experiment1}). These results justify our experimental design in the main paper, where we utilized GT bounding boxes to rigorously evaluate pose estimation performance decoupled from detection errors.

\setcounter{table}{0}
\renewcommand{\thetable}{C\arabic{table}}

\begin{table}[h]
\caption{\textbf{End-to-end 6D object pose estimation using CNOS detections.}}
\label{tab:appendix_experiment1}
\centering
\resizebox{\linewidth}{!}{
\begin{tabular}{l|cc| c| ccc}
\toprule
\multirow{2}{*}{\textbf{Scenario}} &
\multirow{2}{*}{\textbf{Light}} &
\multirow{2}{*}{\textbf{Smoke}} &
\textbf{Detection} &
\multicolumn{3}{c}{\textbf{Pose estimation (ADD-0.3d)}} \\
\cline{4-7}
 &  &  &
\textbf{CNOS} &
\textbf{FoundPose} &
\textbf{GigaPose} &
\textbf{PicoPose} \\
\midrule
\multirow{2}{*}{Sports} & Standard & & 17.85 & 2.12 & 9.38 & 8.68 \\
   & Extreme & & 8.49 & 0.69 & 3.44 & 3.75 \\
\midrule
  \multirow{4}{*}{Maintenance} & Standard & & 41.76 & 22.77 & 33.78 & 41.30 \\
   & Extreme & & 22.97 & 11.67 & 15.32 & 21.13 \\
   & Standard & \CheckmarkBold & 30.86 & 14.43 & 19.94 & 23.36 \\
   & Extreme & \CheckmarkBold & 18.52 & 9.73 & 12.56 & 15.25 \\
\midrule
  \multirow{4}{*}{Emergency} & Standard & & 42.10 & 16.88 & 34.51 & 18.21 \\
   & Extreme & & 16.37 & 4.08 & 8.16 & 1.10 \\
   & Standard & \CheckmarkBold & 35.71 & 12.25 & 28.78 & 31.88 \\
   & Extreme & \CheckmarkBold & 16.39 & 2.41 & 6.17 & 2.33 \\
\bottomrule
\end{tabular}}
\end{table}

\subsection{Instance-level and model-free baselines}

Table~\ref{tab:appendix_experiment5} summarizes the instance-level (GDRNPP~\cite{gdrnpp}) and model-free (OnePose++~\cite{onepose}) methods on the \textit{Tennis} sequence to assess annotation quality and task difficulty. GDRNPP achieves high accuracy across both conditions, serving as a fully-supervised upper bound that validates the reliability of our ground truth annotations. In contrast, the reconstruction-based OnePose++ fails significantly due to rapid motion and frequent occlusions. These results highlight the challenging nature of our dataset and suggest that model-based approaches remain a necessary prerequisite for robustness in this domain.

\begin{table}[h]
    \centering
    \caption{\textbf{Additional baseline results on the \textit{Tennis} sequence.}}
    \label{tab:appendix_experiment5}
    \resizebox{\linewidth}{!}{
        \begin{tabular}{l|c|ccc}
            \toprule
            \textbf{Method} & \textbf{Condition} & \textbf{0.1d} & \textbf{0.2d} & \textbf{0.3d} \\
            \midrule
            \multirow{2}{*}{GDRNPP~\cite{gdrnpp} (Instance-level)} & Standard & 84.96 & 95.06 & 96.50 \\
             & Extreme & 74.15 & 90.14 & 93.64 \\
            \midrule
            \multirow{2}{*}{OnePose++~\cite{onepose} (Model-free)} & Standard & 0.46 & 8.09 & 20.96 \\
             & Extreme & 0.11 & 4.76 & 14.15 \\
            \bottomrule
        \end{tabular}
    }
\end{table}

\section{Extended analysis of 6D pose tracking}

\subsection{Evaluation of GoTrack baseline}

\setcounter{table}{0}
\setcounter{figure}{0}
\renewcommand{\thetable}{D\arabic{table}}
\renewcommand{\thefigure}{D\arabic{figure}}

Table~\ref{tab:appendix_experiment2} presents the evaluation of GoTrack~\cite{gotrack} on the \textit{Sports} scenario. The results indicate that the `Direct' tracking mode suffers significant degradation compared to the per-frame baseline. This aligns with our main findings (Table~\ref{tab:experiment3}), confirming that rapid egocentric motion renders the previous frame's pose unreliable for initialization.

\begin{table}[h]
\caption{\textbf{6D object pose tracking using GoTrack.}}
\label{tab:appendix_experiment2}
\centering
\resizebox{0.8\linewidth}{!}{
\begin{tabular}{l|c| ccc}
\toprule
\multirow{2}{*}{\textbf{Object}} &
\multirow{2}{*}{\textbf{Method}} &
\multicolumn{3}{c}{\textbf{GoTrack(GigaPose)}} \\
\cline{3-5}
 & & \textbf{0.1d} &
     \textbf{0.2d} &
     \textbf{0.3d} \\
\midrule
  \multirow{2}{*}{Pingpong} & Per-frame & 1.42 & 3.51 & 17.33 \\
   & Direct & 0.37 & 0.85 & 4.95 \\
\midrule
  \multirow{2}{*}{Tennis} & Per-frame & 13.81 & 35.17 & 44.29 \\
   & Direct & 7.14 & 9.74 & 11.67 \\
\midrule
  \multirow{2}{*}{Bat} & Per-frame & 14.43 & 29.78 & 46.66 \\
   & Direct & 3.94 & 8.35 & 11.62 \\
\midrule
  \multirow{2}{*}{Golf} & Per-frame & 0.45 & 1.56 & 3.87 \\
   & Direct & 0.48 & 0.61 & 0.66 \\
\midrule
  \multirow{2}{*}{Hockey} & Per-frame & 0.57 & 4.95 & 14.24 \\
   & Direct & 0.53 & 2.30 & 4.95 \\
\bottomrule
\end{tabular}}
\end{table}

\subsection{Evaluation under all light conditions}

Table~\ref{tab:appendix_experiment4} summarizes the 6D object pose tracking results in the sports and emergency scenarios under all light conditions. Our findings indicate that the hybrid tracking approach provided performance gains even in the lower-motion Emergency scenario. Furthermore, analyzing the failure cases under extreme conditions suggests that performance enhancement in low-light environments requires feature restoration pre-processing to be successfully applied before the tracking step.

\begin{table}[h]
\caption{\textbf{6D object pose tracking for GigaPose.}}
\label{tab:appendix_experiment4}
\centering
\resizebox{\linewidth}{!}{
\renewcommand{\arraystretch}{0.75}
\begin{tabular}{l|c|ccc|ccc}
\toprule
\multirow{3}{*}{\textbf{Object}} &
\multirow{3}{*}{\textbf{Method}} &
\multicolumn{6}{c}{\textbf{Lighting condition}} \\
\cline{3-8}
& & \multicolumn{3}{c|}{\textbf{Standard}}
  & \multicolumn{3}{c}{\textbf{Extreme}} \\
\cline{3-5} \cline{6-8}
 & & \textbf{0.1d} & \textbf{0.2d} & \textbf{0.3d}
   & \textbf{0.1d} & \textbf{0.2d} & \textbf{0.3d} \\
\midrule
  \multirow{4}{*}{Pingpong} & Per-frame & 2.17 & 5.36 & 17.84 & 2.77 & 5.94 & 15.15 \\
   & Direct & 0.33 & 1.25 & 5.22 & 0.36 & 1.03 & 2.01 \\
   & Fusion & 0.76 & 2.61 & 11.43 & 0.92 & 2.30 & 8.34 \\
   & Hybrid & 2.44 & 4.59 & 19.07 & 3.32 & 5.84 & 16.51 \\
\midrule
  \multirow{4}{*}{Tennis} & Per-frame & 9.29 & 42.64 & 59.20 & 6.13  & 35.59 & 52.03 \\
   & Direct & 2.92 & 14.52 & 23.64 & 0.57 & 7.79  & 14.84  \\
   & Fusion & 4.98 & 38.12 & 57.49 & 5.56 & 28.37 & 47.62  \\
   & Hybrid & 10.83 & 42.61 & 58.72 & 6.36 & 34.84 & 50.77 \\
\midrule
  \multirow{4}{*}{Bat} & Per-frame & 20.35 & 42.67 & 63.93 & 14.13 & 34.93 & 57.25 \\
   & Direct & 7.55 & 18.20 & 24.57 & 0.0 & 0.72 & 1.92 \\
   & Fusion & 14.28 & 38.95 & 64.92 & 11.36 & 35.47 & 60.82 \\
   & Hybrid & 23.08 & 49.05 & 67.86 & 16.99 & 39.49 & 59.75 \\
\midrule
  \multirow{4}{*}{Golf} & Per-frame & 0.11 & 1.48 & 8.30 & 0.03 & 1.08 & 8.36 \\
   & Direct & 0.63 & 1.86 & 3.92 & 0.0 & 0.50 & 2.33 \\
   & Fusion & 0.66 & 5.53 & 14.07 & 0.65 & 2.10 & 7.13 \\
   & Hybrid & 0.30 & 3.71 & 15.47 & 0.20 & 3.18 & 14.66 \\
\midrule
  \multirow{4}{*}{Hockey} & Per-frame & 0.29 & 4.46 & 16.26 & 0.13 & 2.33 & 7.50 \\
   & Direct & 1.77 & 9.34 & 15.38 & 1.86 & 4.19 & 7.48 \\
   & Fusion & 1.38 & 10.16 & 20.44 & 0.67 & 5.04 & 11.74 \\
   & Hybrid & 0.91 & 8.56 & 22.43 & 0.40 & 3.08 & 10.18  \\
\midrule
  \multirow{4}{*}{Fire extinguisher} & Per-frame & 9.35 & 34.85 & 51.54 & 6.89 & 15.73 & 24.95 \\
   & Direct & 8.75 & 21.12 & 25.55 & 3.78 & 6.62 & 8.21  \\
   & Fusion & 13.45 & 34.25 & 42.79 & 5.36 & 10.43 & 15.72  \\
   & Hybrid & 13.34 & 40.95 & 56.19 & 9.40 & 20.31 & 28.40  \\
\midrule
  \multirow{4}{*}{Kit} & Per-frame &  18.04 & 23.89 & 26.62 & 7.22 & 8.80 & 12.11 \\
   & Direct & 19.31 & 23.31 & 24.52 & 2.60 & 4.38 & 4.83 \\
   & Fusion & 20.92 & 26.38 & 29.16 & 7.96 & 10.44 & 11.62  \\
   & Hybrid & 30.81 & 37.98 & 40.47 & 13.32 & 16.53 & 17.85  \\
\midrule
  \multirow{4}{*}{Flashlight} & Per-frame & 6.37 & 22.71 & 25.19 & 2.43 & 6.67 & 7.52 \\
   & Direct & 6.99 & 16.52 & 17.20 & 0.17 & 1.59 & 1.66 \\
   & Fusion & 8.17 & 20.11 & 21.29 & 0.32 & 2.16  & 2.85  \\
   & Hybrid & 9.72 & 26.55 & 28.77 & 2.54  & 7.01 & 7.68  \\
\bottomrule
\end{tabular}}
\vspace{-3ex}
\end{table}

\subsection{Qualitative results}

\begin{figure}[h]
  \centering
  \includegraphics[width=\linewidth]{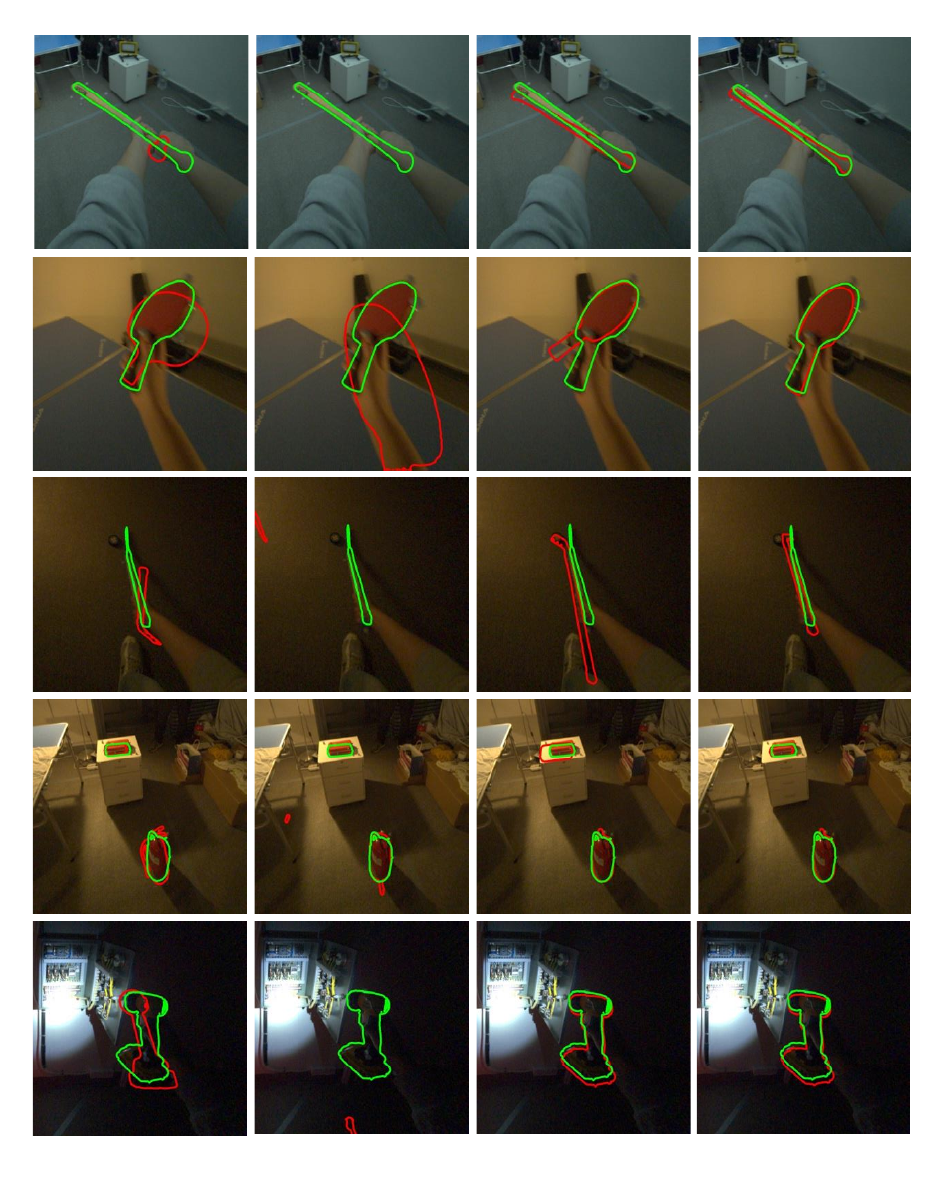}
  \caption{\textbf{Visualization of pose tracking performance.} The panels, arranged from left to right, showcase the results of the per-frame baseline, direct temporal, fusion temporal, and hybrid temporal methods.}
  \label{fig:appendix_qualitative2}
\vspace{-5ex}
\end{figure}

Figure~\ref{fig:appendix_qualitative2} visualizes the comparative performance of four distinct 6D pose temporal strategies. In the high-motion sports scenario, prediction often failed using the simple direct temporal approach due to large inter-frame displacement, leading to frequent failures. While the fusion temporal approach performed better, it still struggled to accurately stabilize rotation. The hybrid temporal method ultimately showed the most significant and reliable improvement in this difficult setting. Conversely, in the emergency scenario, where object movement was relatively minimal, the fusion temporal approach already provided a satisfactory performance level. However, the hybrid method still yielded the greatest overall stability gain, demonstrating its superior ability to fuse reliable measurements over simple propagation.

\section{Extended analysis of image restoration}

\subsection{Pre-processing results under all conditions}

Table~\ref{tab:appendix_experiment3} details the impact of image restoration pre-processing for 6D object pose estimation, analyzing the efficacy across specific illumination conditions and the presence of simulated smoke. Across all tested sub-conditions, we observe a general performance degradation when applying pre-processing, underscoring its inherent limitations for downstream pose estimation tasks.

\setcounter{table}{0}
\setcounter{figure}{0}
\renewcommand{\thetable}{E\arabic{table}}
\renewcommand{\thefigure}{E\arabic{figure}}

\begin{table}[h]
\caption{\textbf{6D object pose estimation with pre-processing under conditions}.}
\label{tab:appendix_experiment3}
\centering
\resizebox{\linewidth}{!}{
\renewcommand{\arraystretch}{0.85}
\begin{tabular}{l|cc|ccc|ccc}
\toprule
\multirow{2}{*}{\textbf{Scenario}} &
\multicolumn{2}{c|}{\textbf{Conditions}} &
\multicolumn{3}{c|}{\textbf{Pre-processing}} &
\multicolumn{3}{c}{\textbf{PicoPose}} \\
\cline{2-3} \cline{4-6} \cline{7-9}
 & \textbf{Light} & \textbf{Smoke} & \textbf{Deblur} & \textbf{Dehaze} & \textbf{Light enhance} &
 \textbf{0.1d} & \textbf{0.2d} & \textbf{0.3d} \\
\midrule
  \multirow{4}{*}{Sports} & \multirow{4}{*}{Standard} & \multirow{4}{*}{} & & & & 3.13 & 9.48 & 24.61 \\
  & & & \CheckmarkBold & & & 2.77 & 9.12 & 24.42 \\
  & & & & & \CheckmarkBold & 3.18 & 9.77 & 23.49 \\
  & & & \CheckmarkBold & & \CheckmarkBold & 2.81 & 8.91 & 23.12 \\
\midrule
  \multirow{4}{*}{Sports} & \multirow{4}{*}{Extreme} & \multirow{4}{*}{} & & & & 1.80 & 6.61 & 17.86 \\
  & & & \CheckmarkBold & & & 2.08 & 6.59 & 15.20 \\
  & & & & & \CheckmarkBold & 1.87 & 6.71 & 17.38 \\
  & & & \CheckmarkBold & & \CheckmarkBold & 1.84 & 5.99 & 14.13 \\
\midrule
  \multirow{5}{*}{Maintenance} & \multirow{5}{*}{Standard} & \multirow{5}{*}{} & & & & 39.27 & 62.42 & 76.84 \\
  & & & \CheckmarkBold & & & 37.72 & 57.40 & 71.21 \\
  & & & & \CheckmarkBold & & 34.15 & 54.47 & 69.60 \\
  & & & & & \CheckmarkBold & 37.62 & 59.07 & 73.18 \\
  & & & \CheckmarkBold & & \CheckmarkBold & 36.08 & 54.69 & 68.14 \\
\midrule
  \multirow{5}{*}{Maintenance} & \multirow{5}{*}{Extreme} & \multirow{5}{*}{} & & & & 26.44 & 47.18 & 64.09 \\
  & & & \CheckmarkBold & & & 26.83 & 45.11 & 60.08 \\
  & & & & \CheckmarkBold & & 22.93 & 43.40 & 60.65 \\
  & & & & & \CheckmarkBold & 22.13 & 41.02 & 57.58 \\
  & & & \CheckmarkBold & & \CheckmarkBold & 22.59 & 39.17 & 54.01 \\
\midrule
  \multirow{5}{*}{Maintenance} & \multirow{5}{*}{Standard} & \multirow{5}{*}{\CheckmarkBold} & & & & 26.37 & 46.05 & 59.87 \\
  & & & \CheckmarkBold & & & 20.63 & 36.69 & 49.94 \\
  & & & & \CheckmarkBold & & 21.52 & 39.54 & 52.94 \\
  & & & & & \CheckmarkBold & 22.95 & 41.02 & 54.88 \\
  & & & \CheckmarkBold & & \CheckmarkBold & 18.28 & 33.86 & 46.69 \\
\midrule
  \multirow{5}{*}{Maintenance} & \multirow{5}{*}{Extreme} & \multirow{5}{*}{\CheckmarkBold} & & & & 20.97 & 38.50 & 52.30 \\
  & & & \CheckmarkBold & & & 20.15 & 35.73 & 48.63 \\
  & & & & \CheckmarkBold & & 17.92 & 33.69 & 47.51 \\
  & & & & & \CheckmarkBold & 17.31 & 32.41 & 46.12 \\
  & & & \CheckmarkBold & & \CheckmarkBold & 17.03 & 31.47 & 44.03 \\
\midrule
  \multirow{5}{*}{Emergency} & \multirow{5}{*}{Standard} & \multirow{5}{*}{} & & & & 22.67 & 59.11 & 67.83 \\
  & & & \CheckmarkBold & & & 22.59 & 57.74 & 66.23 \\
  & & & & \CheckmarkBold & & 5.13 & 25.67 & 43.33 \\
  & & & & & \CheckmarkBold & 20.47 & 57.08 & 65.60 \\
  & & & \CheckmarkBold & & \CheckmarkBold & 21.62 & 55.92 & 64.60 \\
\midrule
  \multirow{5}{*}{Emergency} & \multirow{5}{*}{Extreme} & \multirow{5}{*}{} & & & & 9.18 & 27.59 & 36.23 \\
  & & & \CheckmarkBold & & & 8.62 & 21.74 & 27.59 \\
  & & & & \CheckmarkBold & & 4.65 & 17.26 & 28.42 \\
  & & & & & \CheckmarkBold & 8.62 & 24.46 & 31.87 \\
  & & & \CheckmarkBold & & \CheckmarkBold & 8.41 & 19.85 & 25.62 \\
\midrule
  \multirow{5}{*}{Emergency} & \multirow{5}{*}{Standard} & \multirow{5}{*}{\CheckmarkBold} & & & & 19.66 & 61.35 & 72.82 \\
  & & & \CheckmarkBold & & & 18.19 & 53.38 & 64.84 \\
  & & & & \CheckmarkBold & & 7.27 & 34.49 & 54.20 \\
  & & & & & \CheckmarkBold & 18.38 & 53.60 & 65.04 \\
  & & & \CheckmarkBold & & \CheckmarkBold & 17.75 & 49.82 & 61.83 \\
\midrule
  \multirow{5}{*}{Emergency} & \multirow{5}{*}{Extreme} & \multirow{5}{*}{\CheckmarkBold} & & & & 9.45 & 24.19 & 31.54 \\
  & & & \CheckmarkBold & & & 7.57 & 19.51 & 25.02 \\
  & & & & \CheckmarkBold & & 2.02 & 13.29 & 26.12 \\
  & & & & & \CheckmarkBold & 8.07 & 20.24 & 25.98 \\
  & & & \CheckmarkBold & & \CheckmarkBold & 7.55 & 18.04 & 24.12 \\
\bottomrule
\end{tabular}}
\end{table}

\subsection{Qualitative results}

\begin{figure}[h]
  \centering
  \includegraphics[width=0.8\linewidth]{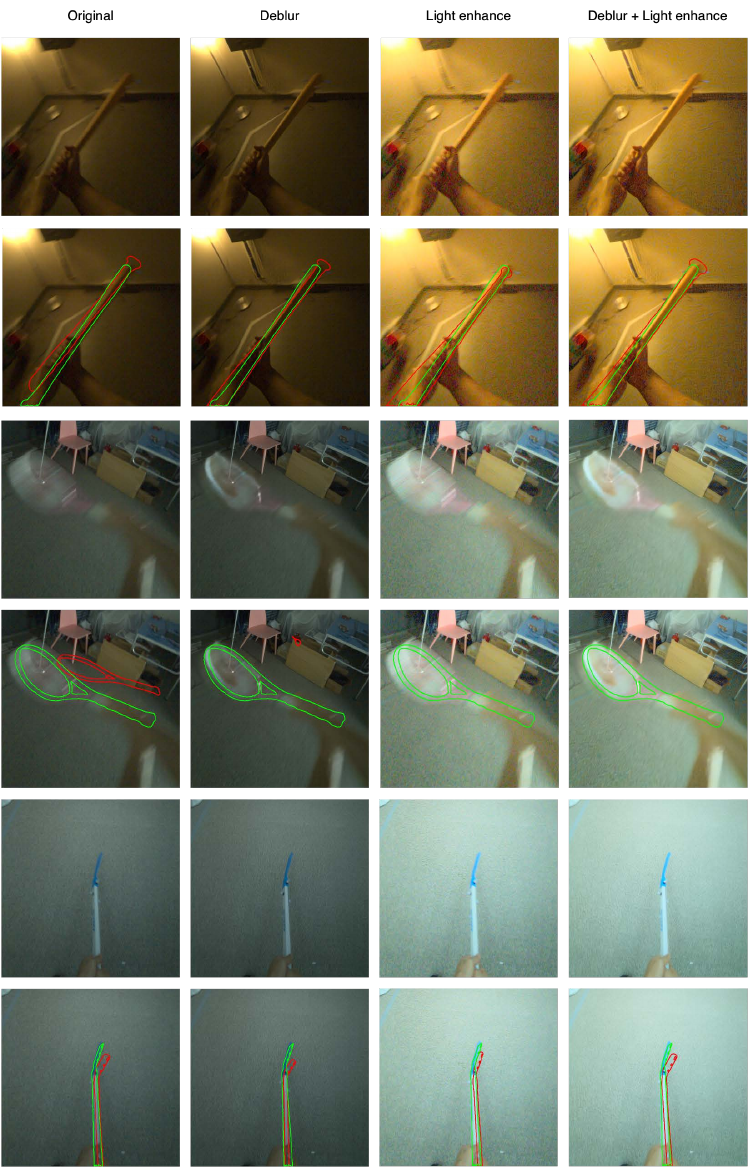}
  \caption{\textbf{Visualization of 6D Pose estimation results with preprocessing on sports scenario.}}
  \label{fig:appendix_qualitative3-1}
\end{figure}

\begin{figure}[h]
  \vspace{-1ex}
  \centering
  \includegraphics[width=0.87\linewidth]{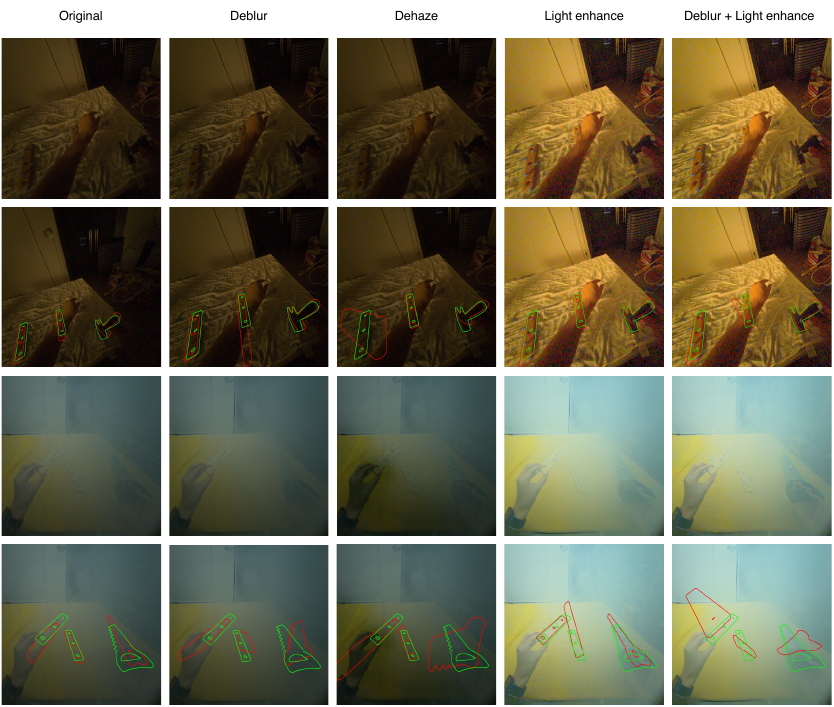}
  \caption{\textbf{Visualization of 6D Pose estimation results with preprocessing on maintenance scenario.}}
  \label{fig:appendix_qualitative3-2}
\end{figure}

\begin{figure}[H]
  \centering
  \includegraphics[width=0.87\linewidth]{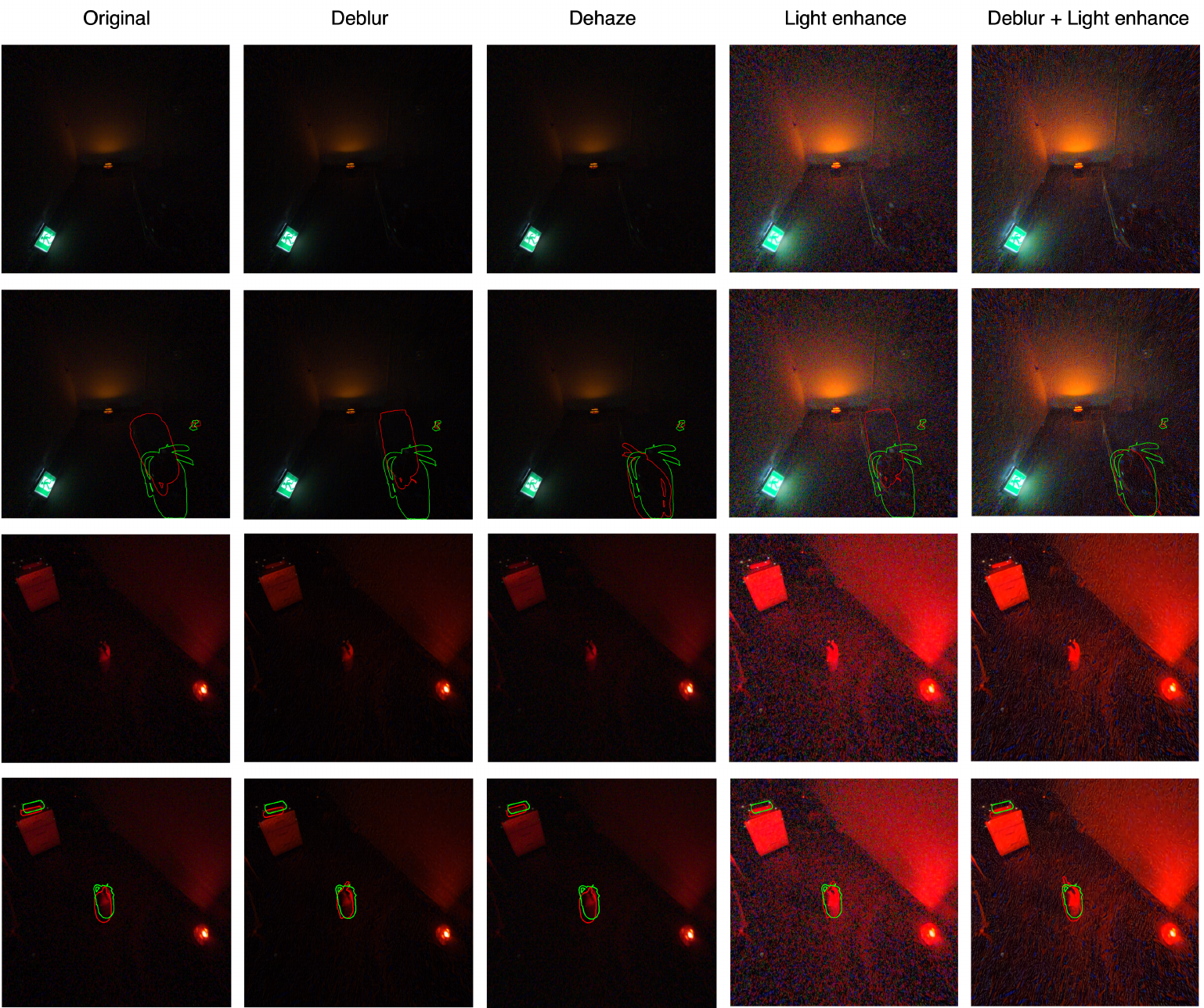}
  \caption{\textbf{Visualization of 6D Pose estimation results with preprocessing on emergency scenario.}}
  \label{fig:appendix_qualitative3-3}
\end{figure}

Figure~\ref{fig:appendix_qualitative3-1},~\ref{fig:appendix_qualitative3-2} and ~\ref{fig:appendix_qualitative3-3} visualizes the impact of image restoration pre-processing on 6D pose estimation. The upper row showcases the visual input across four pre-processing conditions (deblurring, dehazing, light enhancement, and deblurring + light enhancement). The lower row displays the corresponding predicted pose versus the ground truth. Critically, while the processed images show clear visual enhancement, the predicted masks are either not improved or, in some cases, are observed to even worsen.

\clearpage

\section{Institutional Review Board}

The data collection protocol for this study was approved by the Institutional Review Board (IRB) of Seoul National University (IRB No. 2511\_004-017). All participants provided informed consent prior to participating in the study.